\documentclass[letterpaper, 10 pt, conference]{ieeeconf}  

\usepackage{graphicx}

\usepackage{lipsum}

\usepackage[labelformat=simple]{subcaption} 

\usepackage{changepage}
\usepackage{booktabs}

\usepackage{subcaption}
\usepackage{amssymb}
\usepackage{setspace}
\usepackage{multicol, blindtext}
\usepackage{amsmath}
\usepackage{booktabs}
\usepackage{kotex}

\usepackage{array,booktabs,multirow}
\usepackage{url}
\usepackage{color}
\usepackage{epsfig} 
\usepackage{mathtools}

\usepackage{verbatim} 
\usepackage{adjustbox}
\usepackage{enumerate}
\usepackage{threeparttable}
\usepackage[section]{placeins}
\usepackage{graphicx}
\usepackage{caption}
\usepackage{tabularray}
\usepackage{tabularx}
\usepackage{gensymb}
\usepackage{amsfonts}

\makeatletter

\def\env@sqcases{%
  \let\@ifnextchar\new@ifnextchar
  \left\lbrack
  \def\arraystretch{1.2}%
  \array{@{}l@{\quad}l@{}}%
}
\makeatother

\IEEEoverridecommandlockouts   
\newcommand{\RomanNumeralCaps}[1]
    {\MakeUppercase{\romannumeral #1}}

\title{\LARGE \bf
Real-time Whole-body Model Predictive Control for Bipedal Locomotion with a Novel Kino-dynamic Model and Warm-start Method
}

\author{Junhyung Kim$^{1}$, Hokyun Lee$^{1}$ and Jaeheung Park$^{1,2}$
\thanks{*This work was supported by the National Research Foundation of Korea(NRF) grant
funded by the Korea government(MSIT) (No. 2021R1A2C3005914).}
\thanks{$^{1}$Junhyung Kim, Hokyun Lee and Jaeheung Park are with Department of Intelligence and Information, Graduate School of Convergence Science and Technology,  Seoul  National  University,  Republic  of  Korea.
        {\tt\small (john3.16, hkleetony and park73)@snu.ac.kr}}%
\thanks{$^{2}$Jaeheung Park is also with Advanced Institutes of Convergence Technology(AICT), Republic of Korea. He  is  the corresponding author of this paper.
        }%
}

\begin{document}

\maketitle
\thispagestyle{empty}
\pagestyle{empty}

\begin{abstract}
Advancements in optimization solvers and computing power have led to growing interest in applying whole-body model predictive control (WB-MPC) to bipedal robots.
However, the high degrees of freedom and inherent model complexity of bipedal robots pose significant challenges in achieving fast and stable control cycles for real-time performance.
This paper introduces a novel kino-dynamic model and warm-start strategy for real-time WB-MPC in bipedal robots. Our proposed kino-dynamic model combines the linear inverted pendulum plus flywheel and full-body kinematics model. Unlike the conventional whole-body model that rely on the concept of contact wrenches, our model utilizes the zero-moment point (ZMP), reducing baseline computational costs and ensuring consistently low latency during contact state transitions. Additionally, a modularized multi-layer perceptron (MLP) based warm-start strategy is proposed, leveraging a lightweight neural network to provide a good initial guess for each control cycle.
Furthermore, we present a ZMP-based whole-body controller (WBC) that extends the existing WBC for explicitly controlling impulses and ZMP, integrating it into the real-time WB-MPC framework.
Through various comparative experiments, the proposed kino-dynamic model and warm-start strategy have been shown to outperform previous studies. Simulations and real robot experiments further validate that the proposed framework demonstrates robustness to perturbation and satisfies real-time control requirements during walking. 

\end{abstract}

\section{INTRODUCTION}
Model predictive control (MPC) has significantly advanced walking robot technology by enabling stable and adaptive locomotion across various terrains. Based on a dynamics model, The MPC scheme  predicts future states and optimizes control inputs, allowing bipedal robots to respond effectively against perturbations.

Early studies on walking models for MPC utilized simplified linear models such as the linear inverted pendulum model \cite{c6} and linear inverted pendulum plus flywheel model (LIPFM) \cite{c10}. These models assume a point mass at a constant height where the mass of the robot is concentrated, and the LIPFM further assumes that angular momentum is generated at this point mass. Subsequent studies advanced to nonlinear models like the single rigid body model  \cite{c12} and centroidal dynamics \cite{c13}, replaced the zero-moment point (ZMP) with contact wrenches. Although these models improve physical realism, they still do not fully ensure the feasibility of generated walking patterns and reflecting the dynamic effects of whole-body (WB) motion.

Recently, studies have been conducted to overcome the limitations of previous models by controlling or planning bipedal walking robots using whole-body dynamics (WBD) or centroidal dynamics with full-body kinematics model (FKM) \cite{c14},\cite{c16}.
 The use of these WB models in bipedal locomotion improves robustness and feasibility \cite{c25}, \cite{c26}, but incurs substantial computational burdens. Fortunately, advancements in warm-start strategy of whole-body model predictive control (WB-MPC) have led to solve WB-MPC more efficiently. Specifically, \cite{c2} and \cite{c20} introduced warm-start strategies that accelerate dynamic differential programming (DDP) solver by finding a good initial guess from the pre-collected motion datasets.
These memory of motion based warm-start strategies have significantly reduced the computational burden of WB-MPC and enabled the finding of more optimal solutions. However, machine learning-based warm-start strategies remain infeasible in every control cycle. 
Another way to solve WB-MPC more quickly is by limiting the solver's iterations. 
In \cite{c20} and \cite{c5}, the maximum DDP iterations was limited to 1. However, real-time implementation remained impossible due to peak latency during contact transitions, which caused discontinuities in joint torques and failed to outperform state-of-the-art centroidal walking controllers. Another study \cite{c27} implemented WB-MPC using low accuracy solutions obtained within a limited solver iteration, but this approach may cause failures as the stability of solution is not guaranteed.

In this study, we propose a novel kino-dynamic model and warm-start strategy to reduce the computational burden and enable the real-time WB-MPC. 
The proposed model combines LIPFM and FKM, utilizing ZMP instead of contact wrenches, distinguishing it from other WB models. This approach allows for the consideration of dynamic effects of WB motion without requiring inverse dynamics calculations and avoids the discontinuities in the formulation of WB-MPC problem caused by changes in the contact phase.
\begin{figure}[t]
\begin{center}
\includegraphics[width=4.46cm]{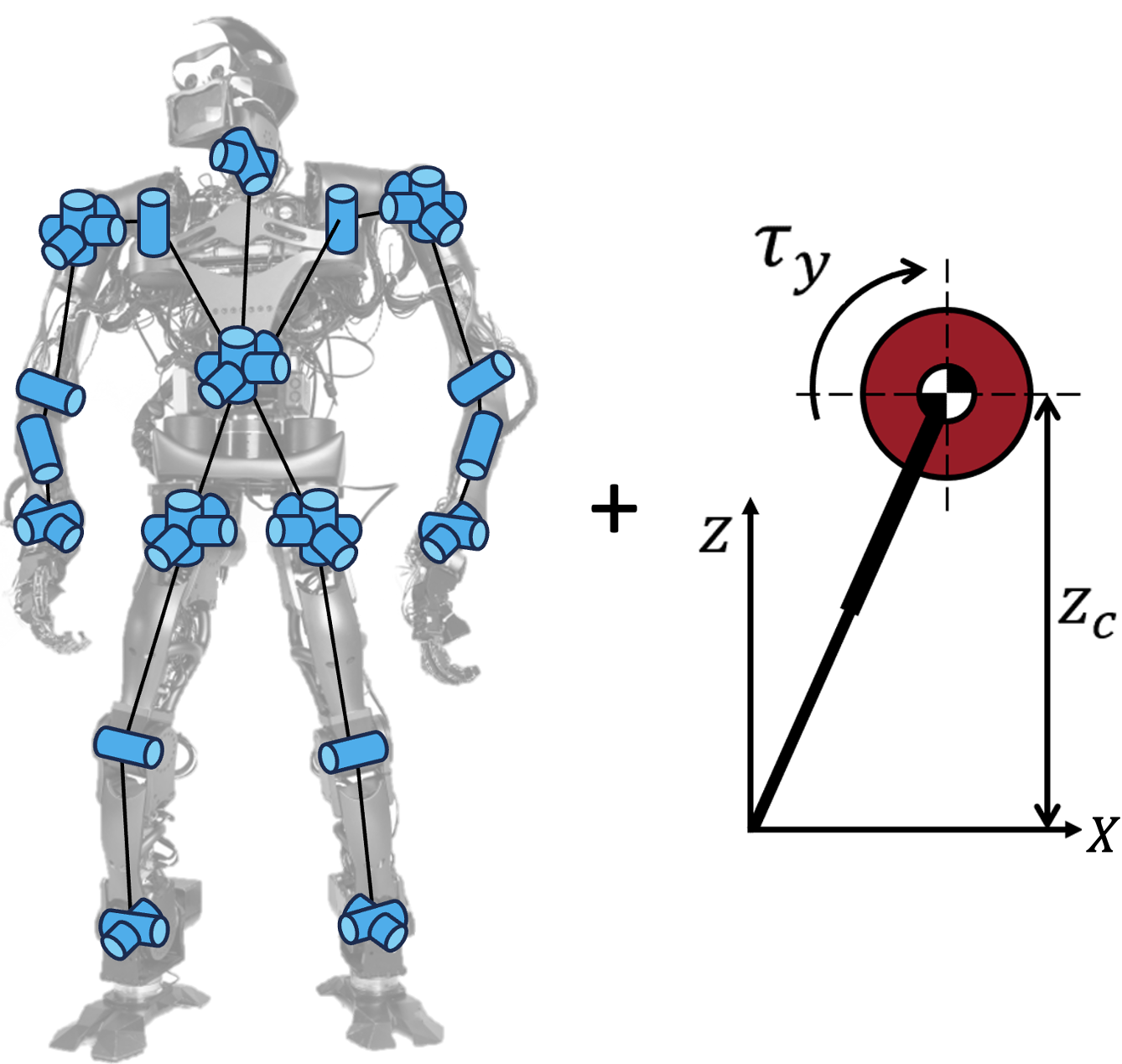}
\caption{Proposed novel kino-dynamic model: Integration of LIPFM and FKM.}
\label{LIPFM}
\vspace{-13pt}
\end{center}
\end{figure}

Additionally, our proposed warm-start strategy utilizes a modularized and lightweight multi-layer perceptron (MLP), unlike existing machine learning-based memory of motion strategy. By modularizing and lightweighting the prediction model, 
our approach achieves a significantly lower latency in finding a good initial guess compared to previous studies.
This enables rapid solving of DDP at each control cycle. All of these ensure real-time control performance and robust stability against perturbation.
To implement MPC effectively, a whole-body controller (WBC) is crucial, as demonstrated in \cite{c3} and \cite{c4}. In this work, we improve the robustness of our WB-MPC framework by introducing a ZMP-based WBC which controls both the ZMP and impulses during walking.

The remainder of this paper is organized as follows. Section \uppercase\expandafter{\romannumeral2} explains a novel kino-dynamic model and proposed control framework for real-time WB-MPC.
Section \uppercase\expandafter{\romannumeral3} describes the modularized MLP based warm-start strategy to solve WB-MPC in real-time. Section \uppercase\expandafter{\romannumeral4} presents the results to verify the performance of the proposed framework. Section \uppercase\expandafter{\romannumeral5} presents the conclusion of this paper.
\section{WHOLE-BODY MPC LOCOMOTION USING PROPOSED NOVEL KINO-DYNAMIC MODEL}

\subsection{LIPFM with Full-Body Kinematics Model}
Previous studies proposed models for reflecting the WB motion, such as WBD or centroidal dynamics with FKM. However, these models have high complexity and cause problem discontinuities during contact state transitions, hindering real-time control performance \cite{c5}.
To address these issues, we propose a novel kino-dynamic model that combines LIPFM and FKM as shown in Fig.\ref{LIPFM}. The proposed model eliminates problem discontinuities during contact state transitions and has relatively low model complexity by replacing the concept of contact wrenches with ZMP.

The LIPFM, is the simplest ZMP-based model capable of reflecting angular momentum, without causing problem discontinuities during contact state transitions.
The equations of the LIPFM are as follows.
\begin{align}
\ddot{x} = \frac{g}{z_{c}}(x-p_{x}) - \frac{\tau_{y}}{m z_{c}} \label{flywheelx}\\\ddot{y} = \frac{g}{z_{c}}(y-p_{y}) + \frac{\tau_{x}}{m z_{c}} \label{flywheely}
\end{align}
where, $x$ and $y$ are the center of  mass (COM) coordinates in the x and y-direction, $p_{x}$ and $p_{y}$ are the ZMP coordinates in the x and y-direction,
$\tau_{x}$ and $\tau_{y}$ are the time derivative of centroidal angular momentum (CAM) in the x and y-direction, $\ddot{x}$ and $\ddot{y}$ are the accelerations of the COM in the x and y-direction, $g$ is the gravity force, $z_{c}$ is the height of the COM, and $m$ is the total mass of the robot, respectively.

The time derivative of CAM generated by WB motion can be expressed from the centroidal momentum matrix as follows.
\begin{equation}
\boldsymbol{H}(\boldsymbol{q})\, \ddot{\boldsymbol{q}} + \dot{\boldsymbol{H}}(\boldsymbol{q})\, \dot{\boldsymbol{q}} = \boldsymbol{\tau} \label{centroidaleq}
\end{equation}
where, $\boldsymbol{H}(\boldsymbol{q})$ and $\dot{\boldsymbol{H}}(\boldsymbol{q})$ are the angular part of centroidal momentum matrix and its time derivative, $\dot{\boldsymbol{q}}$ and $\ddot{\boldsymbol{q}}$ are the velocity and acceleration of joints, and $\boldsymbol{\tau}$ is the time derivative of CAM, respectively. 

Also, kinematic constraints for WB motion can be given through a FKM. The constraints of the COM and placement of end-effectors are determined as follows.
\begin{align}
\boldsymbol{T}_e = \bar{X}_e(\boldsymbol{q})\label{contacteq}\\
\boldsymbol{x}_{com} = \bar{X}_{com}(\boldsymbol{q})\label{comeq}
\end{align}
where,  $\boldsymbol{T}_e$ is the placement of end-effectors and $\bar{X}_c$ is forward kinematic of end-effectors, $\boldsymbol{x}_{com}$ is the position of COM and $\bar{X}_{com}$ is forward kinematic of COM, respectively.

\subsection{WB-MPC Problem Formulation for Locomotion}
The state and control input of the WB-MPC are set as follows.
\begin{equation}
\begin{aligned}
\boldsymbol{x}  = [\overbrace{\boldsymbol{q}, \boldsymbol{\dot{q}},}^{\substack{\text{FKM} \\ \text{variables}}} \overbrace{x, y, \dot{x}, \dot{y}, p_{x}, p_{y}, h_{x}, h_{y}}^{\text{LIPFM variables}}] \\ \boldsymbol{u} =[\overbrace{\boldsymbol{\ddot{q},}}^{\substack{\text{FKM} \\ \text{variables}}} \overbrace{\dot{p}_{x},\dot{p}_{y}, \tau_{x}, \tau_{y}}^{\text{LIPFM variables}}]
\label{state}
\end{aligned}
\end{equation}
where,  $h_{x}$ and $h_{y}$ are CAM in the x and y-direction.

Also, the WB-MPC is formulated as follows.
\begin{figure*}[ht]
\begin{center}
\includegraphics[width=12.5cm]{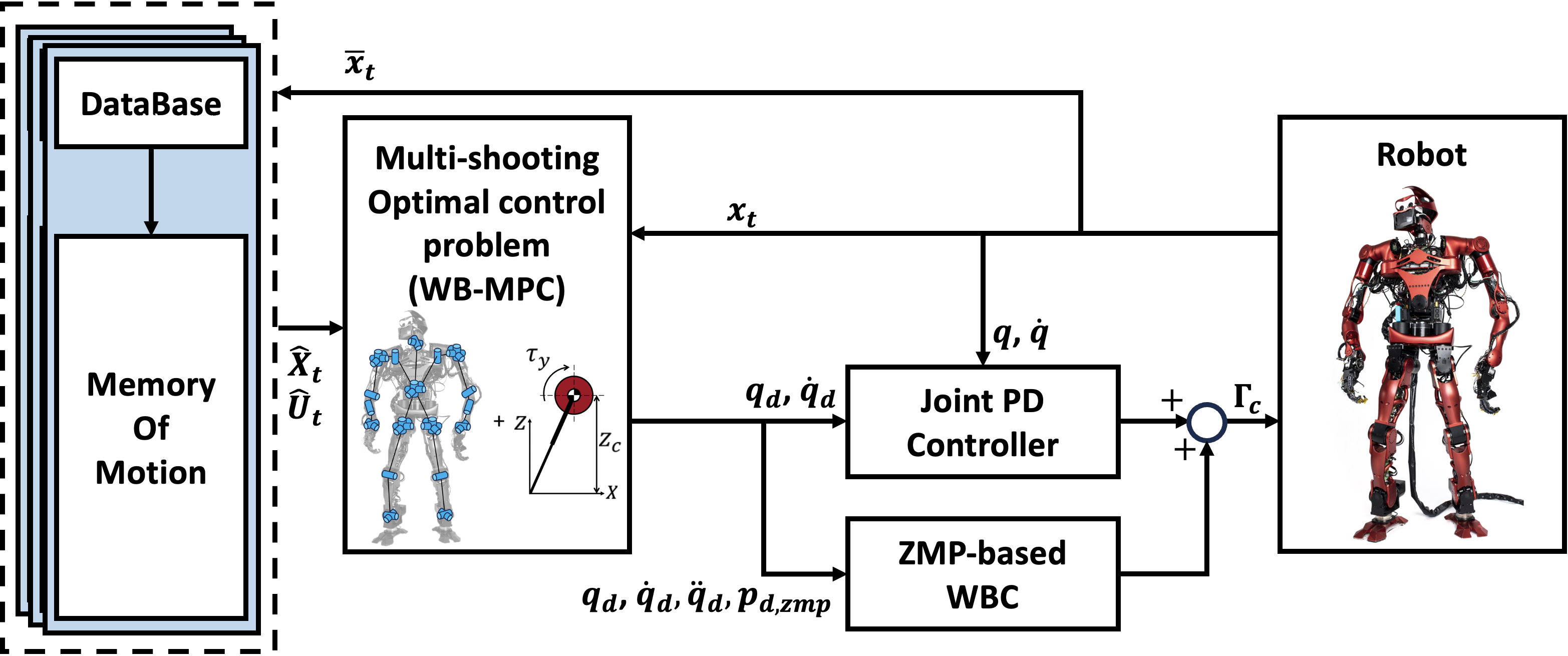}
\caption{The proposed control framework using novel kino-dynamic model.}
\label{controlframework}
\end{center}
\vspace{-17pt}
\end{figure*}
\begin{align}
\displaystyle \min_{\boldsymbol{X_{t},U_{t}}} & \sum_{i=0}^{N-1} \ell(\boldsymbol{x}_{t+i},\boldsymbol{u}_{t+i}) + \ell_{T}(\boldsymbol{x}_{t+N}) \label{dynamicsmodel1}\\
\textrm{s.t.} \quad
&\boldsymbol{x}_{t+i+1} = g(\boldsymbol{x}_{t+i},\boldsymbol{u}_{t+i}) \label{dynamicsmodel}\\  
&h_{1}(\boldsymbol{x}_{t+i}, \boldsymbol{u}_{t+i}) = 0\label{dynamicsmodel1}\\  
  &  \boldsymbol{u}_{min} <\boldsymbol{u}_{t+i} < \boldsymbol{u}_{max} \nonumber
\end{align}
$\boldsymbol{X_{t}} =  [\boldsymbol{x}^{*}_t, \boldsymbol{x}^{*}_{t+1}, \cdot \cdot \cdot,\boldsymbol{x}^{*}_{t+N}]$ and
$\boldsymbol{U_{t}} =  [\boldsymbol{u}^{*}_t, \boldsymbol{u}^{*}_{t+1}, \cdot \cdot \cdot,\boldsymbol{u}^{*}_{t+N-1}]$ are the optimal state and control input trajectories, $g$ denotes the joint kinematic equations relating joint position, velocity, and acceleration and LIPFM dynamics, $h_1$ denotes the kinematic and dynamic constraints of FKM, $\boldsymbol{x}_{t+i}$ and $\boldsymbol{u}_{t+i}$ are the state and control input at time $t+i$, while $\ell$ and $\ell_{T}$ are the running and terminal cost function, respectively. The joint kinematic equations and LIPFM equations \eqref{flywheelx}-\eqref{flywheely} are incorporated into the formulation of \eqref{dynamicsmodel}, while the equations of the FKM \eqref{centroidaleq}-\eqref{comeq} are defined as \eqref{dynamicsmodel1}. The FKM constraints \eqref{dynamicsmodel1} are 
handled efficiently using a penalty method, as employed in \cite{c23}. Also, boundary constraints of control input are set to prevent abrupt motion.

The running and terminal cost function of WB-MPC consists of 5 terms.

\noindent \textbf{(1)  ZMP reference tracking} : 
 \begin{adjustwidth}{1.6em}{0pt} 
  $\ell_1 = ||\boldsymbol{W}_{p}(\boldsymbol{p} - \boldsymbol{p}_{ref})||^2$ is added to control the ZMP. For the stability of the bipedal robot, it is desirable for the ZMP to be close to the center of support polygon. So, the reference of ZMP is defined as follows.
\begin{flalign*}
 \hspace{-1.5em}
  \setlength{\arraycolsep}{1.3pt} 
  \boldsymbol{p}_{ref}=\left\{\begin{array}{ l l }
   (\boldsymbol{c}_{RF}+\boldsymbol{c}_{LF})/2, & \textrm{\hspace{0.8em}if double support phase (DSP)}\\
   \boldsymbol{c}_{RF}, & \textrm{\hspace{-2.9em}if right foot single support phase (SSP)}\\
   \boldsymbol{c}_{LF}, & \textrm{\hspace{0.8em}if left foot SSP}
  \end{array} \nonumber \right.
\end{flalign*}

\noindent where, $\boldsymbol{W}_{p}$ is the positive definite matrix (PDM) of ZMP reference tracking cost, $\boldsymbol{p} = [p_{x}, p_{y}]^{T}$ and $\boldsymbol{p_{ref}} = [p_{x,ref}, p_{y,ref}]^{T}$ are the current and reference of ZMP, $\boldsymbol{c}_{RF} = [c_{RF,x}, c_{RF,y}]^{T}$ and $\boldsymbol{c}_{LF} = [c_{LF,x}, c_{LF,y}]^{T}$ are the desired  contact positions of the right and left feet in the x and y directions, representing the geometric center of the each foot. These positions are determined by the footstep sequence predefined based on gait time and step length within the planning horizon.
Even with a zero-order hold form of reference ZMP,
boundary constraint of control input  limit the changes in
ZMP, resulting in a smoothed desired ZMP trajectory.

\end{adjustwidth}
\noindent \textbf{(2) Boundary constraint of ZMP} : 
  \begin{adjustwidth}{1.6em}{0pt} 
 $\ell_2 = \boldsymbol{W}_{pl}(||\max(\boldsymbol{p}-\boldsymbol{p}_u,0)||^2 + ||\min(\boldsymbol{p}-\boldsymbol{p}_l,0)||^2)$ is added to maintain the ZMP within the supporting polygon. 
To reduce the computational burden, the ZMP boundary constraint is reformulated from an inequality form to a penalization form, and previous studies \cite{c20}, \cite{c5} have shown a high penalty weight effectively avoids infeasible solutions.
where, $\boldsymbol{W}_{pl}$ is the PDM of ZMP boundary constraint, and $\boldsymbol{p}_{u} = [p_{x,u}, p_{y,u}]^{T}$ and
$\boldsymbol{p}_{l} = [p_{x,l}, p_{y,l}]^{T}$ represent the upper and lower bound of the ZMP support polygon.
\end{adjustwidth}
\noindent \textbf{(3)  Capture point(CP) reference tracking} : 
 \begin{adjustwidth}{1.6em}{0pt} 
   $\ell_3 = ||\boldsymbol{W}_{cp}(\boldsymbol{\xi} - \boldsymbol{\xi}_{ref})||^2$  is added for tracking the CP to enhance the  stability of the walking robot.
 The CP reference trajectory is generated from the pre-generated optimal walking trajectory through offline WB-MPC.
 where, $\boldsymbol{W}_{cp}$ is the PDM of CP reference tracking cost, and $\boldsymbol{\xi} = [\xi_{x}, \xi_{y}]^{T}$ and $\boldsymbol{\xi_{ref}} = [\xi_{x,ref}, \xi_{y,ref}]^{T}$ are the current and the reference of CP, respectively.
\end{adjustwidth}

\noindent \textbf{(4)  Feet trajectory tracking} : 
\begin{adjustwidth}{1.6em}{0pt} 
 $\ell_4 = ||\boldsymbol{W}_{e}log(\boldsymbol{T}_{e}^{-1} \boldsymbol{T}_{e,d})||^2$ is added to guide the feet to desired contact position based on predetermined feet trajectories. where, $\boldsymbol{W}_{e}$ is the PDM of feet trajectory tracking, and $\boldsymbol{T}_{e,d}$ is the desired end-effector placements in 
in $SE(3)$, which, in this paper, refers to the feet.
\end{adjustwidth}

\noindent \textbf{(5) Upper body joints regularization :}

\begin{adjustwidth}{1.6em}{0pt} 
Excessive movement of upper body during walking can seem unnatural.
To mitigate this, the cost of regularizing upper body joints is defined as $\ell_5 = ||\boldsymbol{W}_{u}(\boldsymbol{q}_{u} - \boldsymbol{q}_{u,init})||^2$, ensuring that the joints remain close to the initial pose.
where, $\boldsymbol{W}_{u}$ is the PDM for upper body joint regularization, and $\boldsymbol{q}_{u}$ and $\boldsymbol{q}_{u,init}$ are the current and initial position of the upper body joints, respectively.
\end{adjustwidth}

\subsection{ZMP-based Whole-body Controller}
When the robot walks while tracking joint position and velocity trajectories generated by WB-MPC, it is crucial to control the impulse to ensure stable contact of the swing foot with the ground.
In \cite{c3} and \cite{c4}, WBCs were introduced that use relaxations of state variables to generate feed-forward torques that 
control impulses while tracking the desired joint trajectories. These demonstrate improved control performance when combined with MPC. Our proposed model replaces contact wrenches with ZMP, requiring a modification of the existing WBC.
Building on this, we propose a ZMP-based WBC that effectively controls both impulses and ZMP during walking.
\begin{align}
\min_{\delta \ddot{\boldsymbol{q}}, \delta\boldsymbol{F}_{c,z}, \ddot{\boldsymbol{x}}_c,  \boldsymbol{\Gamma}_{wb}} & \ddot{\boldsymbol{x}}_c^{T}W_{c}\ddot{\boldsymbol{x}}_c + \delta \ddot{\boldsymbol{q}}^{T}\boldsymbol{W}_{q}\delta \ddot{\boldsymbol{q}} + \delta \boldsymbol{F}_{c,z}^{T}\boldsymbol{W}_{F}\delta \boldsymbol{F}_{c,z} \nonumber\\
\textrm{s.t.} \quad
  &\mathrm{\boldsymbol{M}\ddot{\boldsymbol{q}}+\boldsymbol{N}(\boldsymbol{q},\dot{\boldsymbol{q}}) = \boldsymbol{J}_c^{T}\boldsymbol{F}_{c}+\begin{bmatrix}
\boldsymbol{0_{6\times1}}\\
\boldsymbol{\Gamma_{wb}}
\end{bmatrix}} \nonumber\\  
  &\textrm{$\ddot{\boldsymbol{q}} = \ddot{\boldsymbol{q}}_d + \delta\ddot{\boldsymbol{q}}$}   \nonumber\\
  &\textrm{$\boldsymbol{J}_c\ddot{\boldsymbol{q}} + \boldsymbol{\dot{J}}_c\dot{\boldsymbol{q}} = \ddot{\boldsymbol{x}}_c$
  }\nonumber\\
    &\textrm{$\boldsymbol{\Gamma}_{min} < \boldsymbol{\Gamma}_{wb} < \boldsymbol{\Gamma}_{max}$}
    \nonumber\\
    &\textrm{$ \boldsymbol{UF}_{c} \leq 0$, \ \ \ $\mathcal{C} \subseteq \left\{ \textnormal{\footnotesize \textit{RF}}, \textnormal{\footnotesize \textit{LF}} \right\} $}\nonumber\\
    &\mathrm{\boldsymbol{p}_{d,zmp} = \frac{\sum_{c_i \in \mathcal{C}} \ \boldsymbol{p}_{c_i,zmp}F_{c_i,z} }{\sum_{c_i\in \mathcal{C}} F_{c_i,z}}} \label{zmp1}\\
    &\boldsymbol{p}_{c_i,zmp} = \frac{\begin{bmatrix} - \tau_{c_iy},  \tau_{c_i,x}\end{bmatrix}^{T}}{{F_{c_i,z}}}  + \begin{bmatrix} c_{c_i,x} \\ c_{c_i,y} \end{bmatrix}\label{zmp2}\\
    &\textrm{$ \delta \boldsymbol{F}_{c,z}=\boldsymbol{F}_{c,z}-\boldsymbol{F}_{c,z}^{ref}$} \label{zmp3} 
\end{align}
where, $\boldsymbol{\Gamma}_{wb}$ is the computed torque commands from WBC, $\boldsymbol{\Gamma}_{max}$ and $\boldsymbol{\Gamma}_{min}$ are the maximum and minimum  limitations of joint torque, $\boldsymbol{J}_c$ and $\boldsymbol{\dot{J}}_c$ are the contact Jacobian and its time derivative, $\boldsymbol{M}$ and  $\boldsymbol{N}(\boldsymbol{q},   \boldsymbol{\dot{q}})$ are the mass matrix and nonlinear terms of robot dynamics, $\boldsymbol{\ddot{x}}_{c}$ is the acceleration of contact point, $\delta\boldsymbol{ \ddot{q}}$ is the relaxation of joint acceleration, $\boldsymbol{U}$ is the matrix which computes normal and friction cone forces, $\boldsymbol{W}_{c}$, $\boldsymbol{W}_{q}$, and $\boldsymbol{W}_{F}$ are the PDMs for regularizing acceleration of contact point, relaxation of joint acceleration, and z-directional contact force, $\mathcal{C}$ is the valid contact set, $\boldsymbol{p}_{d,zmp}$ is the desired ZMP, $\boldsymbol{p}_{c_i,zmp}$ is the local ZMP of $i$-th contact point, $c_{c_i,x}$ and $c_{c_i,y}$ are x and y-directional contact position of $i$-th contact point, $\boldsymbol{F}_{c} = [\boldsymbol{F}_{c,x},\boldsymbol{F}_{c,y},\boldsymbol{F}_{c,z},\boldsymbol{\tau}_{c,x},\boldsymbol{\tau}_{c,y},\boldsymbol{\tau}_{c,z}]$ is the contact wrench of the valid contact set, with $\boldsymbol{F}_{c,z}$ as its z-component, $\tau_{c_i,x}$ and $\tau_{c_i,y}$ are the x and y-directional contact torque of $i$-th contact point, included in $\boldsymbol{\tau}_{c,x}$ and $\boldsymbol{\tau}_{c,y}$, and $\boldsymbol{F}_{c,z}^{ref}$ is the reference of the z-directional contact force for the valid contact set.

The constraints \eqref{zmp1} and \eqref{zmp2} control the contact wrenches by distributing it to achieve the desired ZMP. However, only ZMP constraints cannot control the contact forces needed to reduce the impact when the swing foot lands on the ground. To address this, an additional constraint to control the z-directional contact force is introduced in \eqref{zmp3}.
The reference of z-directional contact force is determined as follows.
\begin{equation}
\begin{aligned}
 \alpha_{c}  =
 \left\{\begin{array}{cc}
    0.5,  & \textrm{if} \  \textrm{right and left foot} \in \mathcal{C},\\
    0.0,  & \textrm{if} \ \textrm{right foot} \in \mathcal{C} \ \textrm{and left foot} \notin \mathcal{C},\\
    1.0, & \textrm{if}  \ \textrm{left foot} \in \mathcal{C} \ \textrm{and right foot} \notin \mathcal{C},\\ 
  \end{array} \right.
  \nonumber
\end{aligned}
\end{equation}

\begin{equation}
\begin{bmatrix}
F_{RF,z}^{ref} & F_{LF,z}^{ref}\end{bmatrix} = \begin{bmatrix}(
   1-\alpha_{c})mg &
   \alpha_{c}mg\end{bmatrix}
   \nonumber
\end{equation}

where, $F_{RF,z}$ and $F_{LF,z}$ are the z-directional contact forces of the right and left feet, and $\alpha_{c}$ is the switching coefficient, respectively. In our proposed algorithm, since the timing of the contact state transition is predefined, $\alpha_{c}$ is interpolated over time with $\delta t = 0.02$ s before the contact state transition to prevent abrupt changes in contact force.

Finally, the command torque is generated by combining the computed torque from WBC with a proportional-derivative controller that tracks the joint position and velocity calculated from WB-MPC, and the overall control framework is shown in Fig. \ref{controlframework}.
\begin{equation}
\boldsymbol{\Gamma}_c = \boldsymbol{\Gamma}_{wb} + \boldsymbol{k}_p(\boldsymbol{q}_{d}-\boldsymbol{q}) + \boldsymbol{k}_v(\dot{\boldsymbol{q}}_{d}-\dot{\boldsymbol{q}})\nonumber
\end{equation}
where, $\boldsymbol{\Gamma}_c$ is joint command torque, $\boldsymbol{k}_p$ and $\boldsymbol{k}_v$ are the proportional and derivative gain of joint controller, $\boldsymbol{q}_{d}$ and $\dot{\boldsymbol{q}}_{d}$ are the desired joint position and velocity, and $\boldsymbol{q}$ and $\dot{\boldsymbol{q}}$ are the current joint position and velocity, respectively.

\section{MODULARIZED MULTI-LAYER PERCEPTRON BASED MEMORY WARM-START}
Solving WB-MPC is challenging and time-consuming due to its inherent complexity and nonlinearity.
To solve WB-MPC more quickly, 
the concept of memory of motion finding a good initial guess from motion databases was proposed in \cite{c2},\cite{c20}. 
Although previous studies demonstrate the effectiveness of memory of motion, applying machine learning algorithms such as Gaussian process regression (GPR) or k-nearest neighbor (k-NN) in real-time has a huge limitation. 
Existing memory of motion strategies rely on the entire training datasets for making predictions, leading to long prediction latency due to their dependence on the size and number of training datasets. This limitation restricts the diversity of the training dataset, making it difficult to utilize diverse motion data in real-time predictions.
In this section, we propose several approaches to address this drawback by efficiently finding good initial guesses from the motion datasets for real-time applications.

\begin{figure}[t]
\begin{center}
\includegraphics[width=7.8cm]{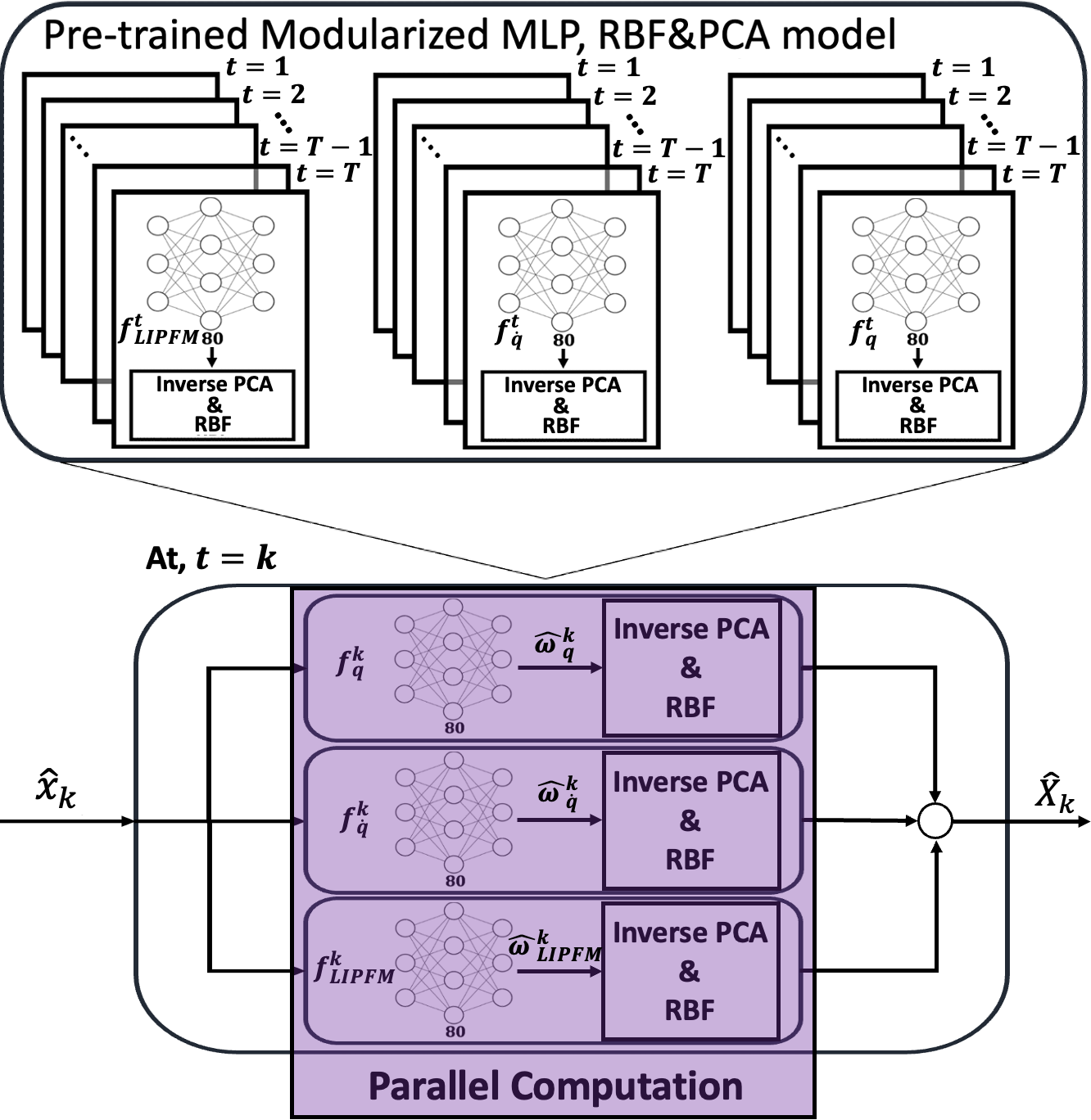}
\caption{Modularized MLP for finding good initial guess}
\label{parallel}
\vspace{-0.8em}
\end{center}

\end{figure}
\vspace{-0.3em}
\subsection{Problem Definition and Training Strategies}
For WB-MPC to quickly find good solutions in each control cycle, it should be able to find a good initial guess based on the current state of robot from motion datasets. Thus, the regression problem can be defined as $f : \boldsymbol{x}_{t} \longrightarrow \boldsymbol{X}_t$ (where, $\boldsymbol{x}_{t}$ and $\boldsymbol{X}_t$ are defined in \eqref{state} and \eqref{dynamicsmodel1}). 
To ensure that a good initial guess is obtained in each control cycle, the initial guess prediction strategy should be lightweight, maintaining low prediction latency to avoid significant impact on the control cycle.
Therefore, reducing the dimensionality of both the input and output of the regression problem is necessary to achieve good performance with a lightweight prediction model. 
The output dimensionality reduction is achieved using radial basis function (RBF) and principal component analysis (PCA), which have been proven effective in previous study \cite{c2}. Consequently, the WB-MPC trajectory $X_t$ is transformed into the weights of RBF $\bar{\boldsymbol{y}}_t = [\omega_1^t , \omega_2^t, ..., \omega_{K}^t] \in \mathbb{R}^{K}$.
To reduce the input dimension of the regression problem, only a subset of the state variables is selected as input. The chosen input $\bar{\boldsymbol{x}}_{t}$ includes the position and velocity of joints, representing kinematic information, and the ZMP, indicating the applied perturbation: $\bar{\boldsymbol{x}}_{t} = [\bold{q_{t}},\bold{\dot{q}_{t}}, p_{x,t}, p_{y,t}]$. Finally, the regression problem becomes $f : \bar{\boldsymbol{x}}_t \longrightarrow \bar{\boldsymbol{y}}_t$.

To train the initial guess prediction strategy, motion datasets are collected for each time step of the walking cycle, as foot placements are predefined according to the trajectories over time. Additionally, to account for perturbation scenarios, we assume that the robot's position and velocity change abruptly due to acceleration caused by the perturbation. So we collect WB-MPC motion datasets by varying the initial joint positions, velocities, and the positions and velocities of the COM and ZMP. The convergence threshold for DDP is set a low value at $10^{-8}$ to generate near-optimal motion trajectory datasets.
\begin{table}[t]
\begin{center}
    \caption{The average latency of model during walking}\label{tab:1}
    \footnotesize{
\begin{tabular}{lc|c|c}
\hline
&Model& Latency/iter & SD\\ 
\hline
&WBD \cite{c5}& 10.2 ms& 2.4 ms\\
&Centroidal Dynamics+FKM \cite{c26}& 9.7 ms& 2.0 ms\\
&LIPFM+FKM & 5.7 ms & 0.2 ms\\
\hline
\end{tabular}
}
\end{center}
\end{table}

\begin{figure}[t]
\begin{center}
\includegraphics[width=6.5cm]{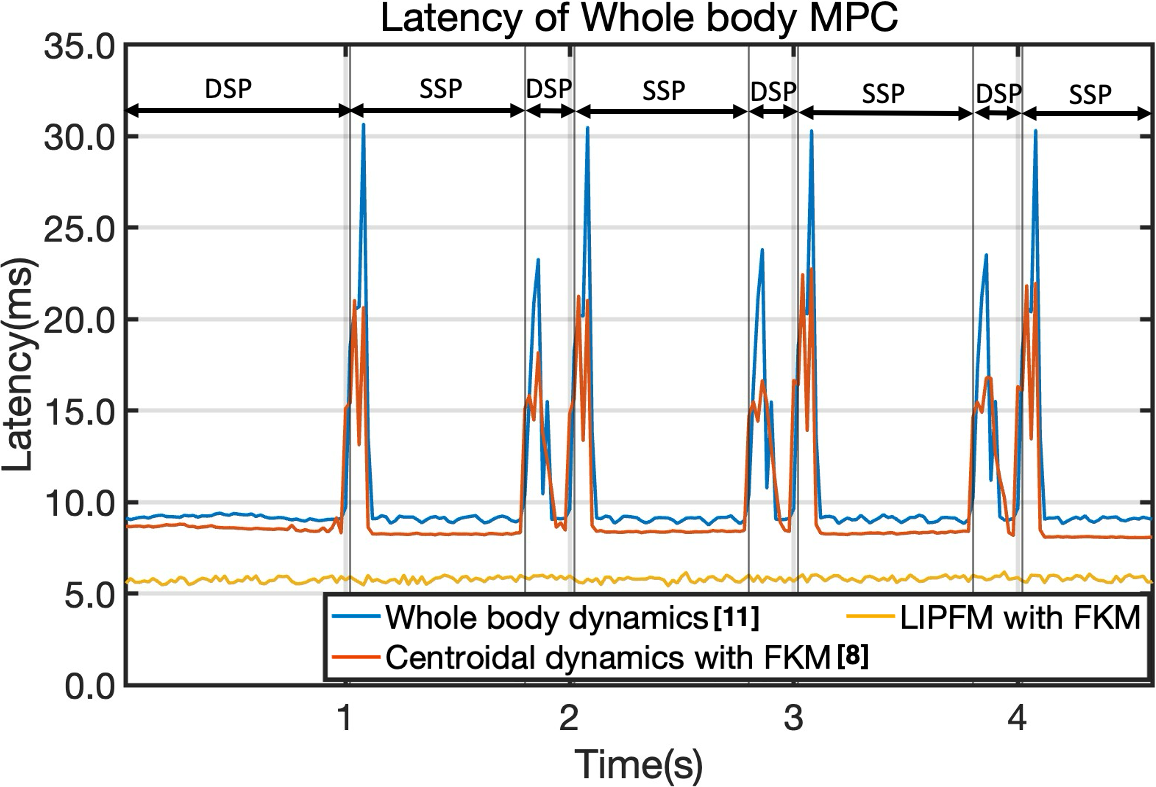}

\caption{Latency graph for 1 DDP iteration during walking}\label{Fig:1}
\label{parallel1}
\end{center}
\vspace{-1.0em}
\end{figure}
\vspace{-0.3em}

\subsection{Modularized MLP for Reducing Prediction Latency}
Two methods are proposed to reduce the initial guess prediction latency compared to previous memory of motion approaches.
First, the prediction model is replaced with a MLP that has a single hidden layer. Unlike GPR or k-NN, the prediction latency of MLP depends on the predetermined number of layers and nodes. Therefore, constructing a lightweight MLP is an effective approach for predicting a good initial guess in each control cycle. 
Second, instead of constructing a single integrated network, the proposed MLP based prediction model is modularly constructed and trained based on the time steps of the walking cycle and the types of state variables (position and velocity of joints, and LIPFM variables), as shown in the following equation and Fig.\ref{parallel}.
\begin{equation}
\label{network}
    \begin{dcases*}
    \bar{\boldsymbol{\omega}}^t_q = f^{t}_{q}(\bar{\boldsymbol{x}}_{t}),\\
    \bar{\boldsymbol{\omega}}^t_{\dot{q}} = f^{t}_{\dot{q}}(\bar{\boldsymbol{x}}_{t}),
    \\
    \bar{\boldsymbol{\omega}}^t_{LIPFM} = f^{t}_{LIPFM}(\bar{\boldsymbol{x}}_{t}),
    \end{dcases*}\nonumber
\end{equation}
where, $\bar{\boldsymbol{\omega}}^{t}_{q}$, $\bar{\boldsymbol{\omega}}^t_{\dot{q}}$, and $\bar{\boldsymbol{\omega}}^t_{LIPFM}$ are the dimensionally reduced WB-MPC data for joint position, joint velocity, and LIPFM state variables at time $t$ through RBF and PCA, and $f^{t}_{q}$, $f^{t}_{\dot{q}}$ and $f^{t}_{LIPFM}$ are the modularized MLP for joint position, joint velocity and LIPFM variables at time $t$, respectively.
This modular approach offers significant benefits: when the model size is limited for lightweighting, its ability to capture data patterns can be restricted, leading to underfitting or reduced accuracy. 
However, segmenting the datasets to train modularized prediction models improves prediction performance.
Furthermore, modularized MLPs enable parallel computation for predicting the initial guess, significantly reducing prediction latency.
\begin{table}[t]
 \begin{center}
    \caption{Prediction Latency and accuracy compared to the machine learning based warm-start strategy}\label{tab:2}
\footnotesize
\begin{tabularx}{\linewidth}{c c c >{\centering\arraybackslash}X >{\centering\arraybackslash}X >{\centering\arraybackslash}X >{\centering\arraybackslash}X}
    \hline
     \multirow{3}{1em}{Error} & & &  \multicolumn{2}{c}{GPR with}  &   \multicolumn{2}{c}{Proposed} \\
     &  & & \multicolumn{2}{c}{RBF\&PCA \cite{c2}} & \multicolumn{2}{c}{Method} \\
    \cline{4-5}    \cline{6-7}
    && & Mean & SD  & Mean & SD\\
    \hline
      & $q$ & &$1.46\!\times\!10^{-5}$&$3.91\!\times\!10^{-5}$&$2.48\!\times\!10^{-5}$&$4.45\!\times\!10^{-5}$\\
& $\dot{q}$ & &$4.04\!\times\!10^{-4}$&$5.25\!\times\!10^{-4}$&$5.12\!\times\!10^{-4}$&$1.08\!\times\!10^{-3}$
\\
 \multirow{2}{0.1em}{LIPFM variables}  & & &\multirow{2}{*}{$8.94\!\times\!10^{-3}$}&\multirow{2}{*}{$1.74\!\times\!10^{-2}$}&\multirow{2}{*}{$0.41\!\times\!10^{-2}$}&\multirow{2}{*}{$1.83\!\times\!10^{-2}$}
\\
\\
  \hline
\multirow{2}{0.1em}{Prediction Latency(ms)} & & & \multirow{2}{*}{528.4}&\multirow{2}{*}{13.15}&\multirow{2}{*}{$\mathbf{0.263}$}&\multirow{2}{*}{$\mathbf{0.195}$}
\\
\\
\hline
\end{tabularx}
\begin{tablenotes}\footnotesize
\item
$\ast$ Training datasets : 8000 samples and Test datasets : 800 samples
\end{tablenotes}
\end{center}
\vspace{-0.8em}
\end{table}
\subsection{Prediction Strategy for Good Initial Guess}
When predicting the initial guess using the pre-trained modularized MLP, the outputs of modularized MLP $\hat{\boldsymbol{\omega}}^t$, $\hat{\boldsymbol{\omega}_q}^t$, and $\hat{\boldsymbol{\omega}}_{\dot{q}^t}$ based on the current state $\boldsymbol{x}_t$ can be transformed into the initial guess of the WB-MPC state trajectory $\hat{\boldsymbol{X}}_t$ by applying inverse transformations of PCA and RBF. Subsequently, the initial guess of the WB-MPC control input trajectory $\hat{\boldsymbol{U}}_t$ can be obtained by solving the dynamics model of WB-MPC \eqref{dynamicsmodel}, using the predicted state trajectory $\hat{\boldsymbol{X}}_t$.
\section{COMPARATIVE ANALYSIS}
This section presents comparative simulation results of the proposed kino-dynamic model and warm-start strategy, focusing on two key objectives: evaluating whether computational latency satisfies the real-time requirements and assessing the effectiveness of the proposed warm-start strategy. 

\subsection{Experiment Setup}
The WB-MPC used in experiments has a horizon of 60 time steps, with a 0.02 s time interval. To solve the WB-MPC, control-limited feasibility-driven DDP \cite{c17} is employed to handle control input constraints.
Experiments are conducted using humanoid TOCABI \cite{c21}. The WB-MPC utilizes 19 actuators: 12 for the legs, 3 for the waist, and 4 for the shoulders.
Other upper body joints are excluded due to the dynamic effects they can produce within the joint constraints are greatly smaller compared to the increased complexity of the problem. Additionally, since motion datasets are collected using only the leg and waist joints, the initial guess generated by the warm-start strategy assumes the shoulder joints are fixed, and comparative experiments are conducted accordingly. All DDP computations runs on Intel i9-13900K with 8 threads, using a convergence threshold of $10^{-5}$ and a max of $5$ DDP iterations to avoid unnecessary computation. Walking simulations are conducted in the MuJoCo physics engine with a 1 s step time, 10 cm step length, and a DSP-to-SSP ratio of 1:4.
The proposed modularized MLP has one hidden layer with 80 LReLUs. To effectively reduce dimension of data, it uses 55 RBFs whose mean and standard deviation (SD) are fine-tuned via line search, and PCA is applied using 32 principal components. 
Finally, the comparison implementation with other WB models followed the problem formulation presented in the referenced papers.
Specifically, \cite{c5} is implemented as a WB-MPC with WBD, which is used for the comparison experiments, and it ensures more stable walking and contact stability than \cite{c2} through a cost term that regularizes the local ZMP.

\begin{table*}[!h]
\begin{subtable}{0.55\textwidth}
    \footnotesize
    \begin{tabular}{cccccc}
        \hline
        \multirow{2}{*}{Error} & \multicolumn{2}{c}{DNN} & \multicolumn{2}{c}{Proposed Method} & Improve \\
        \cline{2-5} 
        & Mean & SD & Mean & SD & ment($\%$)  \\ \hline
        $q$ & $7.89 \times 10^{-5}$ & $5.39 \times 10^{-5}$ & $\mathbf{4.16} \times \mathbf{10^{-5}}$ & $\mathbf{4.37} \times \mathbf{10^{-5}}$ &37.7 \\
        $\dot{q}$ & $6.74 \times 10^{-4}$ & $5.23 \times 10^{-4}$ & $\mathbf{4.53} \times \mathbf{10^{-4}}$ & $\mathbf{5.72} \times \mathbf{10^{-4}}$ &32.5\\
        LIPFM variables & $2.01 \times 10^{-2}$ & $1.71 \times 10^{-2}$ & $\mathbf{2.01} \times \mathbf{10^{-2}}$ & $\mathbf{1.26} \times \mathbf{10^{-2}}$ & 32.3\\
        \hline
    \end{tabular}
    \caption{}
    \label{tab3-a}
\end{subtable}
\hspace{0.13\textwidth}
\begin{subtable}{0.35\textwidth}
    \footnotesize
    \begin{tabular}{cccc}
    \hline
    & \multirow{2}{*}{LSGAN} & \multirow{2}{*}{\begin{tabular}[c]{@{}c@{}}Modularized\\ MLP\end{tabular}} & \multirow{2}{*}{\begin{tabular}[c]{@{}c@{}}Improve\\ ment(\%)\end{tabular}} \\ \\
    \hline
    iter & 1.89 & $\mathbf{1.45}$ & 23.2\\
    cost & 0.0078 & $\mathbf{0.0041}$  & 47.4 \\   
    \hline
\end{tabular}
    \caption{}
    \label{tab3-b}
\end{subtable}
\begin{subtable}{0.55\textwidth}
\footnotesize
\begin{tabular}{cccccc}
        \hline
        \multirow{2}{*}{Error} & \multicolumn{2}{c}{DNN} & \multicolumn{2}{c}{Proposed Method}  & Improve \\
        \cline{2-5} 
        & Mean & SD & Mean & SD &ment($\%$)\\
        \hline
        $q$ & $7.38 \times 10^{-5}$ & $6.70 \times 10^{-5}$ & $\mathbf{5.18} \times \mathbf{10^{-5}}$ & $\mathbf{5.62} \times \mathbf{10^{-5}}$ & 29.8  \\
        $\dot{q}$ & $7.44 \times 10^{-4}$ & $2.74 \times 10^{-4}$ & $\mathbf{5.04} \times \mathbf{10^{-4}}$ & $\mathbf{2.82} \times \mathbf{10^{-4}}$ & 31.8 \\
        LIPFM variables & $2.01 \times 10^{-2}$ & $1.22 \times 10^{-2}$ & $\mathbf{1.49} \times \mathbf{10^{-2}}$ & $\mathbf{1.27} \times \mathbf{10^{-2}}$ & 25.9\\
        \hline
    \end{tabular}
    \caption{}
    \label{tab3-c}
\end{subtable}
\hspace{0.13\textwidth}
\begin{subtable}{0.25\textwidth}
\footnotesize
\begin{tabular}{cccc}
    \hline
    & \multirow{2}{*}{LSGAN} & \multirow{2}{*}{\begin{tabular}[c]{@{}c@{}}Modularized\\ MLP\end{tabular}} & \multirow{2}{*}{\begin{tabular}[c]{@{}c@{}}Improve\\ ment(\%)\end{tabular}} \\ \\
    \hline
    iter & 2.27 & $\mathbf{1.56}$&31.2 \\
    cost & 0.0112 & $\mathbf{0.0058}$ &48.2  \\                                  
    \hline
\end{tabular}
\caption{}
    \label{tab:3-d}
\end{subtable}
\begin{tablenotes}[flushLeft]
\footnotesize
\item
\hspace{0.1cm} $\ast$ Training datasets: 65000 samples and Test datasets: 6000 samples
\end{tablenotes}
\centering
\caption{Comparison good initial guess prediction performance with neural network based algorithm. (a) and (b) are tested on motion data at the start of DSP, while (c) and (d) are tested on motion data at the start of SSP.}
\vspace{-12pt}
\end{table*}

\subsection{Comparing Computational Latency Among WB Model}
To compare the computational latency, WBD and centroidal dynamics with FKM are used as reference models.
Also, simulations are conducted after finishing the WB-MPC computation to ensure comparisons under same control cycle.

In Table \ref{tab:1}, our proposed model shows about a 44 \% reduction in computational latency compared to other WB models. Moreover, it exhibits a significantly smaller SD compared to the other models.
As shown in Fig. \ref{Fig:1}, the computational latency of the reference models increases dramatically—nearly threefold—during the contact state transition, This is attributed to structural changes in the problem caused by contact dynamics, which also lead to abrupt changes in contact wrenches as noted in \cite{c5}. In contrast, our proposed model uses ZMP instead of contact wrenches, maintaining a consistent problem structure even during contact state transitions, thereby sustaining low computational latency. Furthermore, by avoiding inverse dynamics calculations, it achieves approximately a 38 \% lower average computational latency, even during phases without contact transitions.
The primary advantage of the proposed model is its relatively low computational latency and reduced SD compared to other WB models, making it suitable for fast and real-time control.

\subsection{Comparing Warm-Start Strategy}
In this subsection, 
we evaluate the performance of the proposed strategy by comparing the accuracy of the predicted initial guess and prediction latency against previously studied memory of motion strategy and neural network based algorithms with different structures.
Additionally, we assess the performance under perturbation to demonstrate the advantages of warm-start with a good initial guess.

\subsubsection{Comparing with machine learning based strategy}

Table \ref{tab:2} compares the performance difference between the proposed modularized MLP with a machine learning based warm-start strategy \cite{c2} in terms of prediction latency and accuracy. Although the prediction accuracy of machine learning based strategy is slightly higher than that of the proposed strategy, the proposed strategy achieves warm-start with less than 1 ms latency, compared to approximately 500 ms with GPR.
This is because GPR prediction latency is influenced by the size of the training dataset. In our study, we used motion datasets that include perturbation scenarios, making the dataset much larger than in previous work \cite{c2} and resulting in increased latency.
  Consequently, relying on the existing machine learning-based strategy for real-time applications limits the size and diversity of the training dataset. In contrast, the proposed strategy ensures low prediction latency, making it suitable for real-time performance while allowing diverse training datasets.
\subsubsection{Comparing with neural network based algorithms}
To validate the superiority of the proposed strategy, performance comparisons are conducted by training WB-MPC motion data directly on a deep neural network (DNN) without dimensionality reduction and modularization by types of state variables and walking time steps, and implementing least squares generative adversarial networks (LSGAN) to compare with unsupervised learning strategies. For the experiment, the DNN consists of 3 hidden layers with 120, 420, and 1200 LReLU, respectively. The LSGAN is trained so that the discriminator distinguishes between the motion trajectory generated by the generator and the real motion trajectory. The generator and discriminator consist of two hidden layers with 420 and 1200 LReLU. The motion datasets used for comparison are those collected at the start of DSP and SSP.

In Subtable \RomanNumeralCaps{3}-a and \RomanNumeralCaps{3}-c, 
the modularized MLP demonstrates significantly higher accuracy in a predicting good initial guess for all types of state variables compared to the DNN in both DSP and SSP. 
The comparison experiments are trained on data for only two walking time steps, but when a size-limited DNN is trained on data from more walking time steps, the prediction accuracy for a good initial guess at specific walking time steps inevitably decreases.
Furthermore, applying dimensionality reduction to the WB-MPC motion data using PCA and RBF results in R-squared values of 0.99975 and 0.99953, respectively, indicating that the PCA and RBF models effectively capture the nonlinear patterns of motion datasets. 
Therefore, our strategy of training MLP individually based on walking time steps and types of state variables, combined with reducing the dimension of datasets with PCA and RBF, is more effective for a network with limited size.
Additionally, as shown in Subtable \RomanNumeralCaps{3}-b and \RomanNumeralCaps{3}-d, warm-start with modularized MLP achieves a more optimal solution with fewer DDP iterations than LSGAN in both DSP and SSP. These results suggest that, due to the characteristic of optimization problems where the optimal solution is clearly defined for given inputs, supervised learning proves to be more effective than unsupervised learning.  In conclusion, our proposed strategy is demonstrated to be more effective than other neural network approaches.
\begin{figure}[t]
\begin{center}
\includegraphics[width=6.5cm]{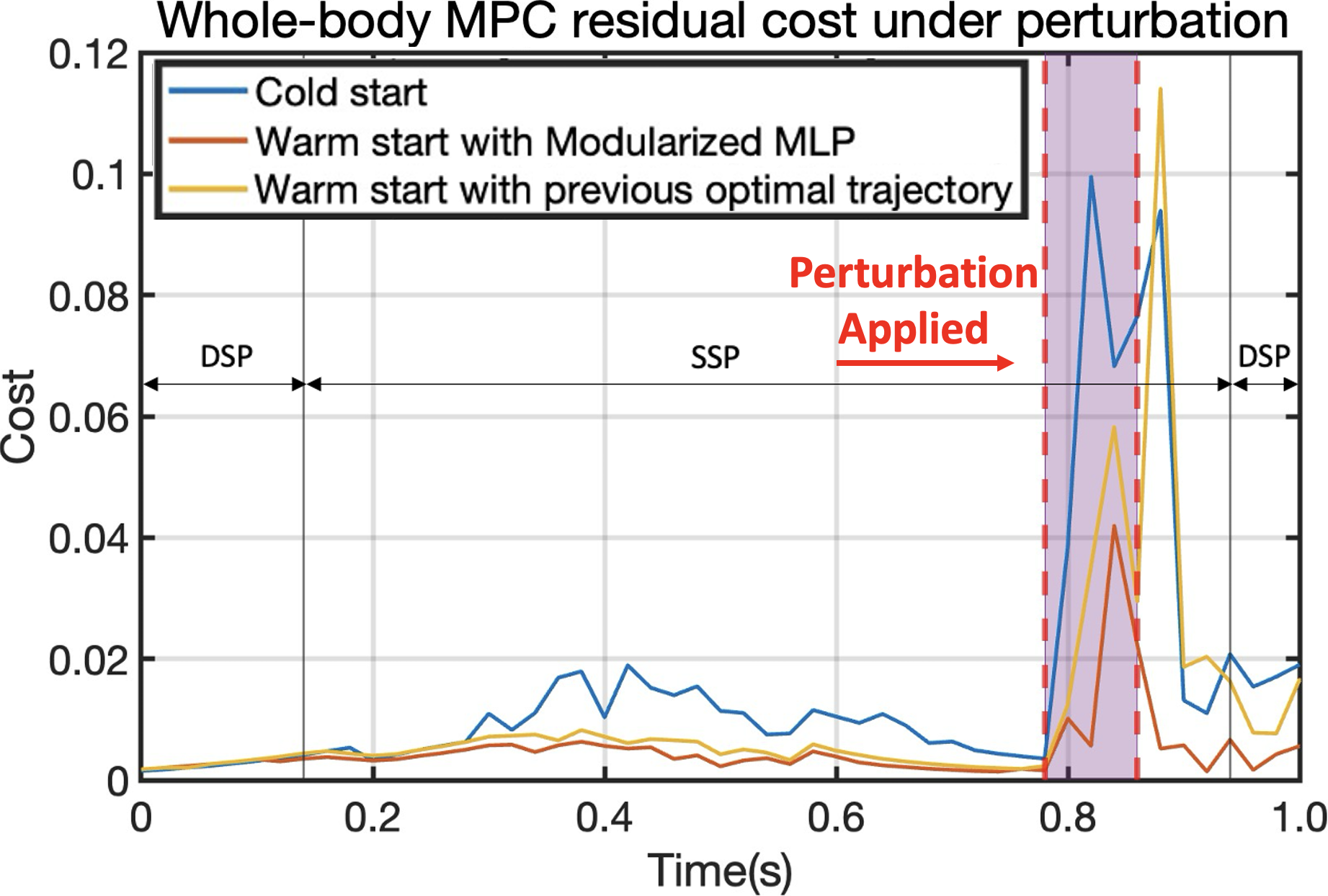}
\caption{The residual cost of WB-MPC under perturbation}\label{fig:5}
\label{parallel2}
\end{center}
\vspace{-15pt}
\end{figure}
\begin{figure}[t]  
    \centering
    \begin{subfigure}{0.295\textwidth}  
        \centering
        \includegraphics[width=\textwidth]{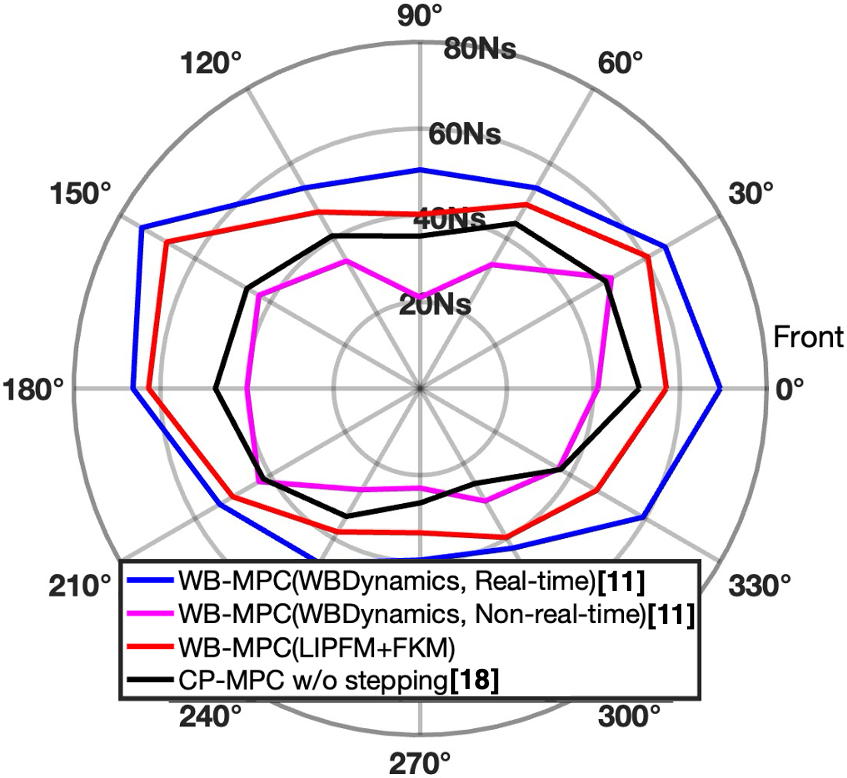}  
        \subcaption{}  
        \label{dis:1}
    \end{subfigure}%
    \hfill  
    \begin{subfigure}{0.18\textwidth}  
    \centering
    \resizebox{0.84\textwidth}{!}{%
    \begin{tabular}{|l|c|c|}
\hline
 & avg & max \\ \hline
iter. & 2.04 & 3 \\ \hline
\multirow{2}{*}{lat. (ms)} & \multirow{2}{*}{9.04} &  \multirow{2}{*}{16.06} \\ 
 &  & \\ \hline
\end{tabular}
    }
    \subcaption{}  
    \label{dis:2}
\end{subfigure}
    \begin{tablenotes}\footnotesize
    \item
$\ast$ "iter." referes to "iteration" and "lat." referes to "latency"
\end{tablenotes}
    \caption{(a) The maximum impulse force a robot can withstand based on the direction of the external force (0\degree  and 90\degree, with perturbation applied to the front and left, respectively). (b) DDP performance under perturbation in real-time simulation.}
\end{figure}

\subsubsection{Comparison under perturbation}
To evaluate the robustness of the proposed warm-start strategy under perturbation, a 40 Ns forward perturbation is applied to the robot in simulation just before the SSP-to-DSP transition (0.78 s and 0.84 s), as shown in Fig. \ref{fig:5}. 
Before the perturbation,  the MPC residual cost is lowest for the warm-start with the modularized MLP, followed by the warm-start with the previous optimal trajectory and cold start, although their values are relatively similar.
However, a significant difference emerges once the perturbation is applied. 
When the perturbation is applied, the MPC residual cost inevitably increases as the ZMP adjusts to maintain balance.
Nevertheless, the proposed warm-start strategy yields a more optimal solution during the perturbation, with the MPC residual cost quickly converging back to pre-perturbation levels even after the perturbation ends, compared to other strategies.
This improvement is attributed to the modularized MLP, which provides a good initial guess from the current state of robot, enabling faster convergence and finding more optimal solutions.

\section{WALKING SIMULATIONS AND EXPERIMENTS}
This section demonstrates the robustness of WB-MPC locomotion using our proposed kino-dynamic model and warm-start strategy by applying perturbation during forward walking in both real-time simulations and real robot experiments. The parameters for walking, DDP, and neural network remain consistent with those used in the previous section \RomanNumeralCaps{4}.

\begin{figure}[t]
    \centering
    \begin{subfigure}{0.5\textwidth}
        \centering
        \includegraphics[width=0.89\linewidth]{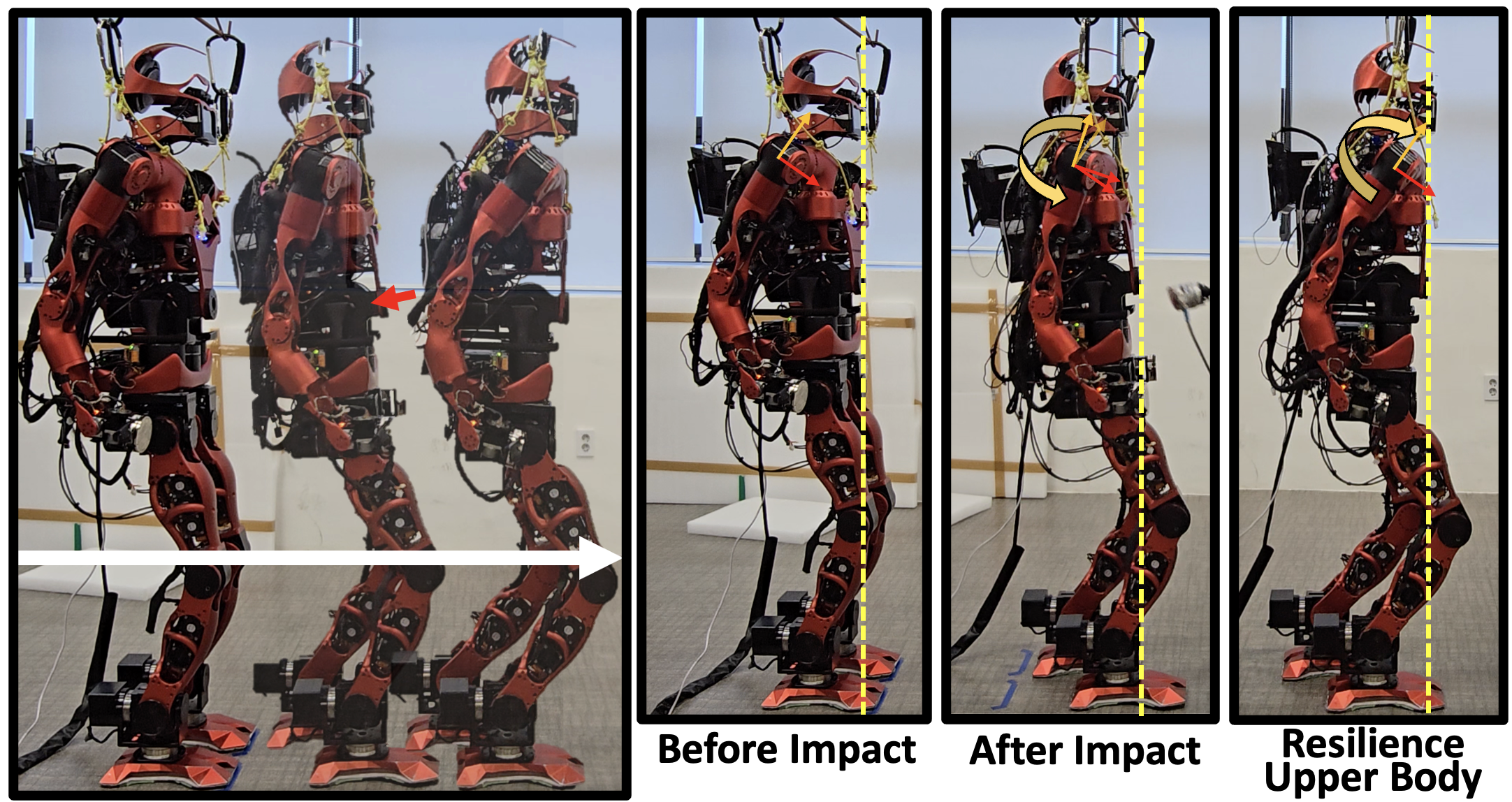}
        \subcaption{}
        \label{fig:parallel5}
    \end{subfigure}
    \begin{subfigure}{0.5\textwidth}
        \centering
        \includegraphics[width=0.95 \linewidth]{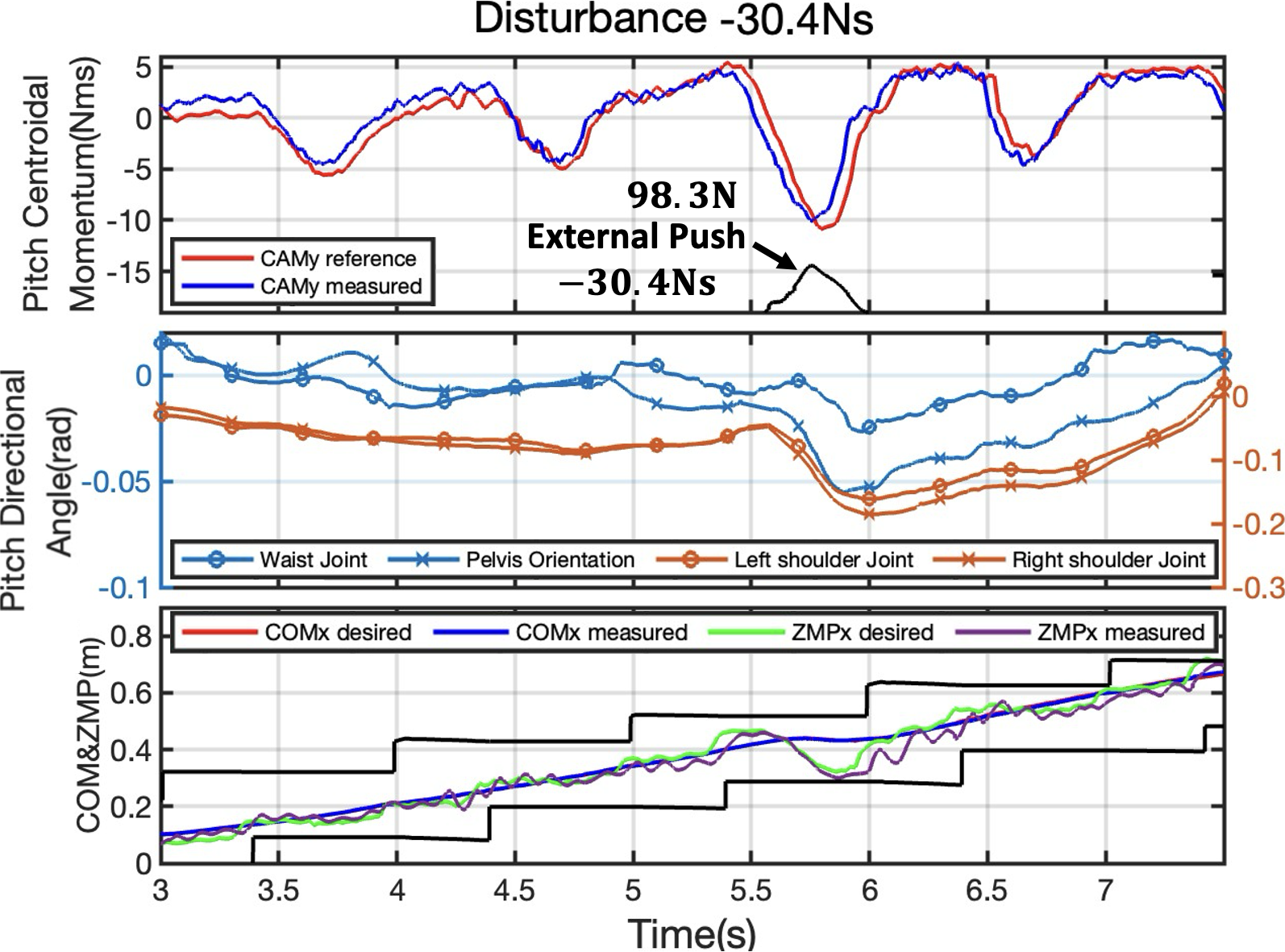}
        \subcaption{}
        \label{fig:parallel6}
    \end{subfigure}
    \caption{(a) Snapshot of walking experiment under negative x-directional perturbation. (b) The walking trajectory and upper body motion generated through WB-MPC under negative x-directional perturbation.}
    \vspace{-2mm}
\end{figure}
\vspace{-2mm}
\subsection{Walking Simulation under Perturbation}
Simulations demonstrate the robustness of the proposed WB-MPC by applying step-input perturbation in various directions to the pelvis during SSP-to-DSP transition. To compare its robustness against perturbation, comparative simulations are conducted using WB-MPC with WBD \cite{c5} and the state-of-the-art centroidal walking controller, Capture Point MPC (CP-MPC) \cite{c22}. For a fair comparison, the stepping strategy of CP-MPC, which adjusts foot placement in response to perturbations, is excluded, while all other components were maintained. This is because the proposed method follows a pre-defined foot sequence. Simulations for WB-MPC with WBD are conducted under two conditions: 
one assuming real-time computation within the control cycle and the other with a non-real-time system \cite{c5}, where computation latency occasionally causes delays exceeding 30 ms and violating the control cycle.
As shown in Fig. \ref{dis:1}, in the most ideal case when real-time computation can be performed within 20 ms, WB-MPC with WBD can withstand larger perturbation compared to the proposed WB-MPC. This is because WBD has no assumptions for model simplification. However, when operating in non-real-time, the computation latency of WB-MPC causes a lag in desired torque updates, leading to a loss of balance with smaller perturbation compared to the proposed WB-MPC.
Also, our proposed WB-MPC outperforms CP-MPC in maintaining balance more robustly against perturbation in all directions. This is because the WB model fully accounts for the dynamic effects of WB motions, unlike simplified models.
Furthermore, Fig. \ref{dis:2} shows that the proposed WB-MPC maintains low DDP latency under 16.06 $\text{ms}$ and finds solutions in 2.04 iterations on average, even under perturbation. Lastly, walking stability is maintained even on uneven and soft terrain, as shown in the attached video.

\subsection{Real Robot Experiments under Perturbation}
Real robot experiments with TOCABI, conducted under the same conditions as simulations, validate the robustness of framework and its consistent performance. In the real robot experiments, the robot can withstand negative x-directional impacts of 30.4 Ns and positive y-directional impacts of 15.3 Ns.
As shown in Fig. \ref{fig:parallel5} and \ref{fig:parallel6}, the proposed WB-MPC framework maintains balance under perturbation by  reactively controlling WB motions, including pelvis orientation, upper body joints, as well as the COM, ZMP, and CAM.
When a perturbation is applied to the robot at 5.7 $s$, a maximum pelvis pitch angle of -0.058 rad, a waist pitch joint angle of -0.027 rad, and shoulder joint angles of -0.18 rad are utilized to maintain balance. After achieving stability, the robot recovers its original posture within 2 seconds through upper body joint regularization costs. Additionally, during 12 real-world robot experiments under perturbation, the proposed framework demonstrates stable performance by maintaining a low DDP latency of under 16.83 ms and finding solutions within an average of 2.35 iterations.
A more encouraging aspect is that, although shoulder joints are not considered in the motion datasets for predicting the initial guess, there are no issues with warm-starting the WB-MPC.
Even when the robot configuration and step length differ from the motion datasets used to train modularized MLP, the robot validate stable walking with WB-MPC, proving the robustness of proposed framework even in unseen states.

\section{CONCLUSIONS}
This paper presents a real-time WB-MPC framework for bipedal robot locomotion, featuring a novel kino-dynamic model and an efficient warm-start strategy. The proposed kino-dynamic model combines the a ZMP-based dynamics model with FKM, significantly reducing computational burden and resolve peak latency issues during contact state transitions compared to the previous WB model. This ensures the real-time performance of WB-MPC across all walking phases. Furthermore, the proposed modularized MLP based warm-start strategy demonstrates superior performance over existing memory of motion algorithms by providing a good initial guess at every control cycle, significantly leveraging real-time applicability.
We also introduced a ZMP-based WBC, specifically adapted to complement for our WB-MPC framework. Through various experiments, the computational efficiency and robustness of our WB-MPC framework have been validated. Even under perturbation, we can solve the WB-MPC with 60 time steps, 4620 variables, 2400 constraints, and various penalty-form constraints within a latency of 17 ms, while the robot maintains balance. 

However, the proposed kino-dynamic model, based on the ZMP concept with vertical COM constraints, has limitations in ensuring stability on terrains requiring vertical COM motion, such as stairs or steep slopes, and cannot account for the flight phase, limiting its ability to generate dynamic motions like running or jumping. Future work will focus on extending the model to overcome these limitations.

\addtolength{\textheight}{-12cm}   





\end{document}